\documentclass[final	]{cvpr}

\usepackage{times}
\usepackage{epsfig}
\usepackage{graphicx}
\usepackage{amsmath}
\usepackage{amssymb}
\usepackage{multirow}
\usepackage{tabularx,booktabs}

\usepackage[pagebackref=true,breaklinks=true,colorlinks,bookmarks=false]{hyperref}

\begin{document}


\title{GuidedStyle: Attribute Knowledge Guided Style Manipulation for Semantic Face Editing}

\author{Xianxu Hou \quad Xiaokang Zhang \quad Linlin Shen \quad Zhihui Lai \quad Jun Wan  \\
Shenzhen University
}

\maketitle

\begin{abstract}
Although significant progress has been made in synthesizing high-quality and visually realistic face images by unconditional Generative Adversarial Networks (GANs), there still lacks of control over the generation process in order to achieve semantic face editing. In addition, it remains very challenging to maintain other face information untouched while editing the target attributes. In this paper, we propose a novel learning framework, called GuidedStyle, to achieve semantic face editing on StyleGAN by guiding the image generation process with a knowledge network. Furthermore, we allow an attention mechanism in StyleGAN generator to adaptively select a single layer for style manipulation. As a result, our method is able to perform disentangled and controllable edits along various attributes, including smiling, eyeglasses, gender, mustache and hair color. Both qualitative and quantitative results demonstrate the superiority of our method over other competing methods for semantic face editing. Moreover, we show that our model can be also applied to different types of real and artistic face editing, demonstrating strong generalization ability.

\end{abstract}

\section{Introduction}
In the past few years there has been significant progress in Generative Adversarial Networks (GANs), the quality of images produced by GANs has improved rapidly. The current state of the art GANs \cite{karras2018progressive,karras2019style,karnewar2020msg,karras2020analyzing} can produce high-fidelity face images at a much higher resolution. However, it remains very challenging to control the generation process of GANs with adjustable semantic specifications. For example, how can we generate or edit a face image with pre-defined attributes like smiling, eyeglasses or mustache?

Some pioneering approaches \cite{shen2020interpreting,abdal2019image2stylegan} that support above mentioned semantic controls have been developed. They tried to achieve face editing by exploring the semantics in the latent space of well-trained GANs like StyleGAN \cite{karras2019style,karras2020analyzing}. However, these models need to obtain the attribute information of the synthesized faces in advance. In such a workflow, the semantic editing relies on the availability of the semantic labels, which might be very difficult to obtain for a synthesized dataset. In addition, existing methods often regard face attribute editing as finding the corresponding linear path (represented as vectors) in the latent space, and then edit different attributes by moving the latent codes along the discovered directions. However, due to the entanglement between different semantics in the latent space, performing edits along one attribute could lead to unexpected changes of other semantics.


\begin{figure}[!]
\begin{center}
   \includegraphics[width=0.98\linewidth]{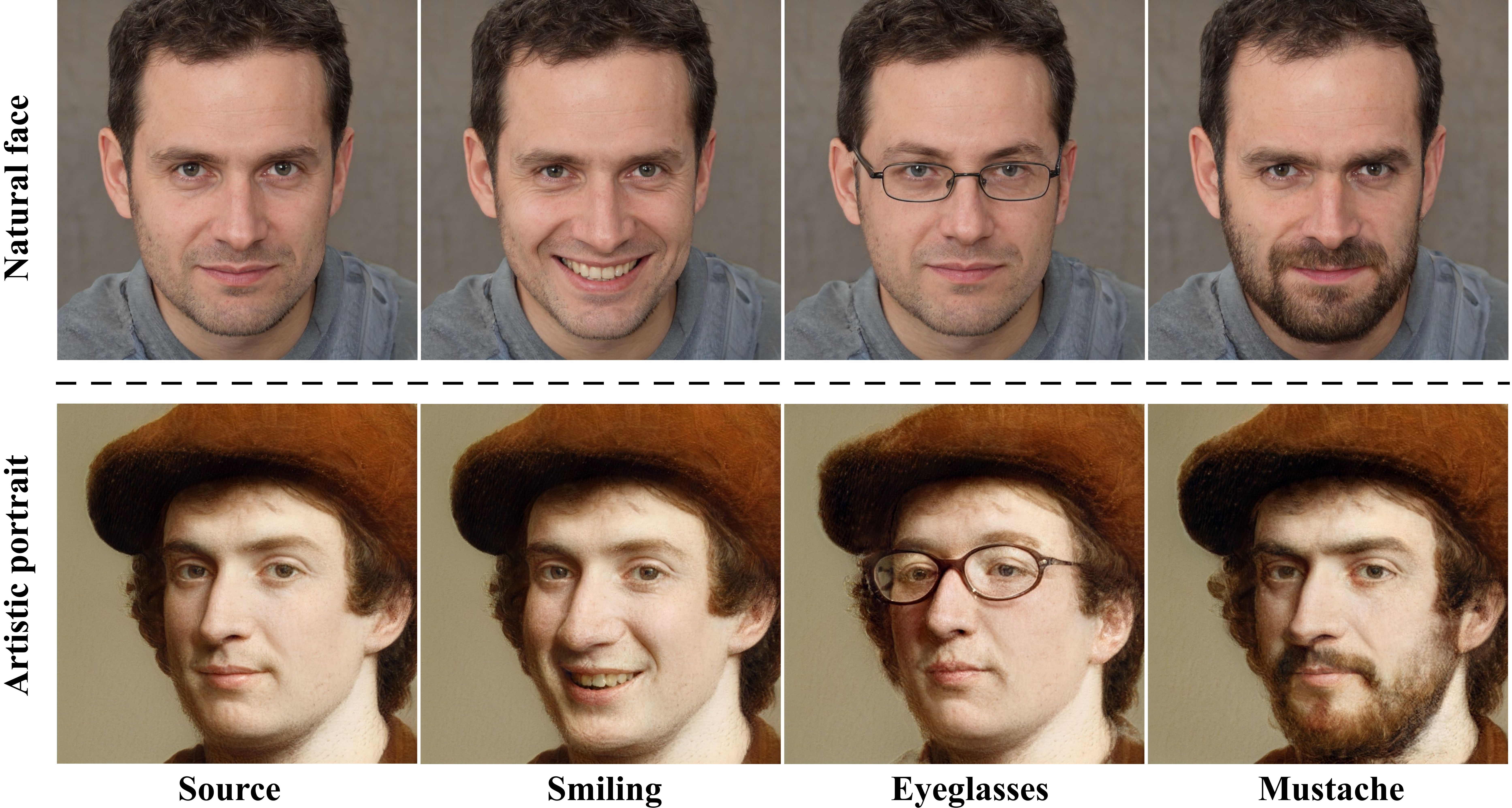}
\end{center}
   \caption{Semantic editing of natural faces (top row) and artistic portraits (bottom row) along different attributes (zoom in for better resolution).}
\label{fig:abstract}
\end{figure}

\begin{figure}[!]
\begin{center}
   \includegraphics[width=1.0\linewidth]{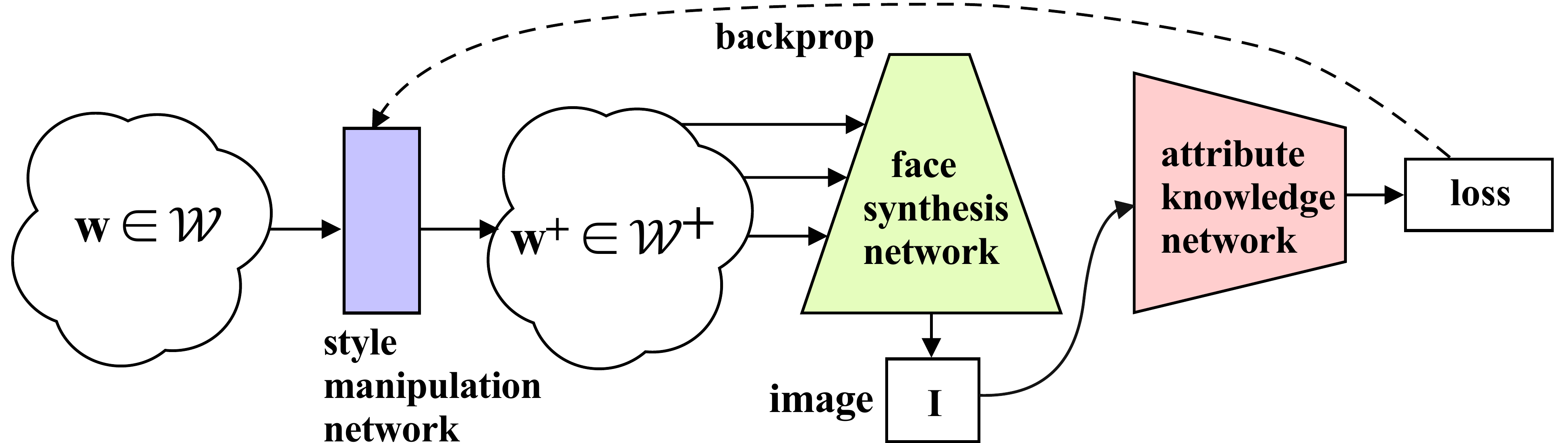}
\end{center}
   \caption{Overview of our method.}
\label{fig:system}
\end{figure}

With the above limitation of existing methods in mind, we argue that there is no need to restrict the editing along the predefined linear path, as what we want is just the correct change of the target attributes. To this end, we propose to learn a non-linear style manipulation for face editing in the latent space of StyleGAN guided by a \textit{knowledge network}. More specifically, a pretrained model for face attribute prediction is used as the knowledge network and provides the supervision signals to learn the correct edits in a \textit{manipulation network} (see Figure \ref{fig:system}). In this way, we are able to ensure the correct edits of the target attributes by using the semantic knowledge learned by the pretrained attribute classifier. Furthermore, we  allow a sparse attention mechanism to adaptively select a single style layer in StyleGAN to perform the semantic manipulation. As a result, the attention mechanism can bring a more disentangled and controllable edits along various attributes such as smiling, eyeglasses, gender, mustache and hair color (see Figure \ref{fig:abstract}). Concretely, we summarize the main contributions of this work as follows:


\begin{itemize}
   \item We present a novel framework, termed as GuidedStyle, to guide the face generation of StyleGAN with the knowledge learned from face attribute classifiers. Our method successfully unifies a face generative model and a face attribute recognition model in a joint framework.

   \item We further incorporate a sparse attention mechanism in the proposed GuidedStyle framework, based on which we can achieve semantic face editing by focusing on a single style layer in the StyleGAN generator.

   \item We validate the effectiveness of our approach on various face editing experiments, demonstrating the superiority of our GuidedStyle framework over other state of the art techniques. Our method can also produce high quality edits on real faces and artistic portraits, demonstrating strong generalization capability.

\end{itemize}

\section{Related Work}
\textbf{Generative Adversarial Networks.}
GANs \cite{goodfellow2014generative} have been at the forefront of research in deep generative models during the past few years, and they can synthesize realistic face images that are almost indistinguishable from real data. However, GANs are notorious for its difficult training and mode collapse. A lot of efforts, including the design of better network architectures \cite{radford2015unsupervised,denton2015deep,zhang2019self,karras2018progressive}, objective functions \cite{mao2017least,arjovsky2017wasserstein,gulrajani2017improved} and training strategies \cite{miyato2018spectral,salimans2016improved,karnewar2020msg}, have been made to improve the training of GANs. In particular, StyleGAN \cite{karras2019style} and StyleGAN2 \cite{karras2020analyzing} are the current state of the art methods in unconditional generative modeling for high-resolution image synthesis. In this work, we build our method on StyleGAN2 as it can synthesize high-quality face images with unmatched photorealism from randomly sampled codes.

\textbf{Face Editing with Conditional GANs.}
Conditional GANs \cite{mirza2014conditional} can be used to achieve image editing by incorporating additional information as input. In the context of faces, a common approach is to use face images and semantic labels as conditional information, then the face editing can be formulated as an image-to-image translation task \cite{isola2017image,zhu2017unpaired}. StarGAN \cite{choi2018stargan,choi2020stargan} proposes a multi-domain image translation framework and transfers different facial attributes by only using a single model. To avoid the irrelevant attribute information, ResidualGAN \cite{shen2017learning} tries to learn residual images to edit different attributes. Fader Networks \cite{lample2017fader} seek to disentangle the salient and attribute information in the latent space, and achieve face editing by varying the attribute values. AttGAN \cite{he2019attgan} edits several attributes by imposing constraint on the translated face images. SwitchGAN \cite{zhu2019switchgan} uses feature switching operation to achieve multi-domain face image translation. Besides editing common facial attributes like gender, smiling and eyeglasses, specialized face editing techniques such as makeup transfer (PairedCycleGAN \cite{chang2018pairedcyclegan} and PSGAN \cite{jiang2020psgan}) and even facial caricature translation (CariGANs \cite{cao2018carigans} and WarpGAN \cite{shi2019warpgan}) are also available. While conditional GANs could provide a certain level of attribute control via image-to-image translation, the edited faces cannot match the resolution and quality of images produced by unconditional GANs.

\begin{figure*}[!t]
\begin{center}
   \includegraphics[width=0.92\linewidth]{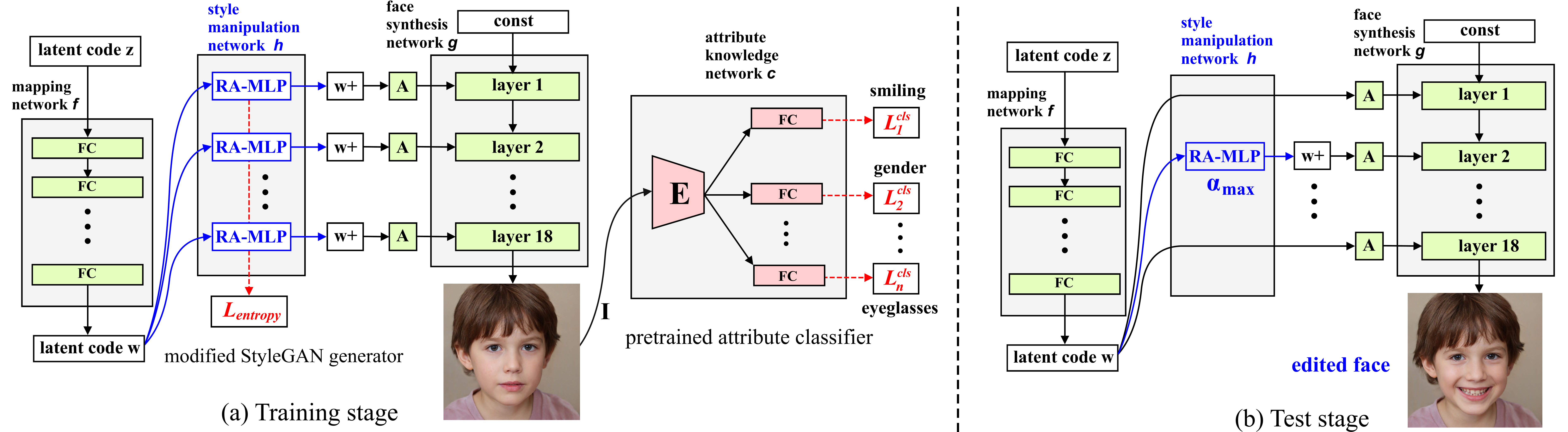}
\end{center}
   \caption{We modify the generator by inserting a style manipulation network with multiple RA-MLP layers between the mapping network and face synthesis network in StyleGAN2. We train the modified generator with the loss functions defined by using a knowledge network pretrained for face attribute prediction. During training stage, only the manipulation network in \textbf{\textcolor{blue}{blue}} is trainable and \textbf{\textcolor{red}{red}} dashed arrows indicate the supervisions. During test stage, we only retain a single RA-MLP layer selected by the maximal attention $\alpha_{\text{max}}$ to perform the edit.
}
\label{fig:overview}
\end{figure*}

\textbf{Face Editing with Pretrained GANs.}
Another successful line of research on semantic face editing is the idea to manipulate the latent space of a pretrained GAN generator. This idea is first adopted by DCGAN \cite{radford2015unsupervised}, which observes that the trained generator have interesting vector arithmetic properties allowing for easy editing of different visual concepts. VAE-GAN \cite{hou2019improving,larsen2016autoencoding} combines VAEs \cite{kingma2013auto,hou2017deep} and GANs \cite{goodfellow2014generative} into a joint generative model, and also demonstrates the vector arithmetic property in the learned latent space. More recently, this approach becomes very popular and several face editing techniques have been proposed with the advancement of StyleGAN. InterfaceGAN \cite{shen2020interpreting} interprets the latent face representation and enables semantic face editing along various attributes by using a linear editing path. StyleRig \cite{tewari2020stylerig} proposes a combination of computer graphic techniques with deep generative models to achieve face rig-like control over a pretrained StyleGAN model. GANSpace \cite{harkonen2020ganspace} considers unsupervised identification of interpretable controls over image synthesis, showing that semantically meaningful directions can be found by applying PCA in the latent space of StyleGAN. However, GANSpace requires extensive effort to manually pick the semantic edits. Similarly, SeFa \cite{shen2020closed} proposes a closed-form factorization method for latent semantic discovery in an unsupervised manner. In addition, a model rewriting approach \cite{bau2020rewriting} is introduced to manipulate specific rules of pretrained GANs and achieves various face edits by modifying the generator. Different from previous works \cite{shen2020interpreting,hou2019improving,harkonen2020ganspace,shen2020closed} that adopt linear editing in latent space with a predefined editing direction, our method seeks to control face attributes via a non-linear editing conditioned on the starting latent code. Another noticeable difference is that our method can adaptively select a single style layer to perform the edits while prior arts are designed to manipulate all the layers or a specific subset selected manually. One concurrent work, called StyleFlow \cite{abdal2020styleflow}, also tries to enable semantic face editing by extracting non-linear paths in the latent space. However, it requires the attribute labels of the synthesized faces to learn a conditional continuous normalizing flow. By contrast, our non-linear mapping is supervised by a pretrained attribute recognition model and is not limited to the generated facial semantics on a particular dataset.

\textbf{GAN Inversion.}
In order to support real image editing with pretrained GANs, a common practice, also known as GAN Inversion, is used to inversely embed images into the latent space of a GAN model. Existing methods either try to learn an encoder network that projects an input image to the corresponding latent code \cite{zhu2016generative}, or directly perform optimization over a latent code based on reconstruction loss functions \cite{creswell2018inverting}. Additionally, other methods adopt a two-stage inversion by combining the two techniques \cite{bau2019seeing,bau2019semantic}, and use the encoder to provide a better starting point for the subsequent optimization. Recently, more and more inversion methods have been designed to embed images into the StyleGAN latent space. Image2StyleGAN \cite{abdal2019image2stylegan} successfully maps an input image into the latent space, and the inversion performance can be improved with additional noise optimization \cite{abdal2020image2stylegan++}. In-domain inversion \cite{zhu2020indomain} tries to keep the inverted code to be semantically meaningful rather than only considering per-pixel reconstruction. In addition, ramped-down noise can be also added to the latent code during optimization \cite{karras2020analyzing}. In this work, GAN inversion serves as a tool to obtain corresponding latent code for real face editing.

\section{Method}
\textbf{Overview.}
We aim to achieve high-quality semantic face editing with well-trained GANs. To achieve this goal, we seek to build a non-linear mapping in the latent space of StyleGAN and further ensure the learned mapping to be semantically meaningful via an attribute knowledge network. Figure \ref{fig:overview} shows the overall flow of our pipeline and the details of our method are provided in the following subsections.

\subsection{StyleGAN based Face Synthesis}
We consider the state of the art generative model, \ie, StyleGAN2 \cite{karras2020analyzing}, which has been very successful in synthesizing high-quality faces by employing a style-based generator. We modify the generator by inserting a style \textit{manipulation network} ($h$), between the \textit{mapping network} and \textit{face synthesis network} shown in the left part of Figure \ref{fig:overview}(a) (illustrated in \textbf{\textcolor{blue}{blue}}). Concretely, we add multiple residual attention multi-layer perceptrons (RA-MLPs) (Figure \ref{fig:ra-mlp}) to further transform latent code $w$ to an extended version $w^+$ before applying it for generation. All the RA-MLP layers are independent of each other and control the generator through adaptive instance normalization \cite{huang2017arbitrary} at each convolution layer of the synthesis network. In this work, we use the modified StyleGAN2 generator for face synthesis.

\subsection{Attribute Knowledge Guided Style Manipulation}

\textbf{Main Idea.}
Recall that our goal is to edit a facial attribute in the latent space of GANs via a non-linear manipulation network. To achieve this goal, a semantic-aware supervision is needed for the training. Our strategy is to make full use of the knowledge learned from the pretrained attribute classifier to guide the image generation process. Intuitively, we would like to let the well-trained classifier \textit{think} that the attribute editing is along the \textit{right} direction. An important insight of our work is that we are able to unify a face generative model and a face attribute recognition model in a joint framework. As a consequence, our approach can support flexible controls over various facial attributes.


\subsubsection{Attribute Knowledge Network}
In our GuidedStyle framework, a pretrained model is required in advance to estimate different face attributes. In this work, we employ a multi-task learning (MTL) model to estimate different attributes simultaneously. In the context of deep learning, MTL typically consists of a shared encoder and several independent fully connected layers to predict different attributes (illustrated in the right part of Figure \ref{fig:overview}(a)). Thus, the multi-attribute estimation is actually a multi-label binary classification task, and the classifier can be trained by using a sigmoid cross-entropy loss function.

\subsubsection{Attribute Knowledge Guided Supervision}
As shown in Figure \ref{fig:overview}(a), the overall flow of the training pipeline is as follows. We first feed the input latent code $z$ to the mapping network $f$ to get the intermediate latent code $w=f(z)$, which will be further transformed to multiple extended latent codes $w^+ = h(w)$ by the newly designed style manipulation network $h$. We then apply the synthesis network $g$ to generate corresponding face image $I=g(w^+)$ with an affine transformation $A$. To enable conditional control over the facial attributes, we feed the synthesized faces to a knowledge network $c$ to construct the loss function. Formally, we build a binary cross-entropy loss $\mathcal{L}^{cls}_i$ when editing the $i^{th}$ attribute as follows:
\begin{equation}
\small
\mathcal{L}^{cls}_i(f,h,g,c) = -y_i\cdot\log(c_i(I)) - (1-y_i)\cdot\log(1-c_i(I))
\end{equation}
where $c_i(I)$ is the predicted score of a particular attribute for the generated image $I$. $y_i$ is the predefined target label, and $y_i = 1$ if we want to add the corresponding attribute, otherwise $y_i = 0$. Thus, the training process can be formulated as
\begin{equation}
h^* = \arg\min_h\mathcal{L}^{cls}_i(f,h,g,c)
\end{equation}
Note that only the inserted style manipulation network $h$ is trainable while all the other components $f$, $g$ and $c$ retain the pretrained weights and keep fixed during the training process.







\begin{figure}[!]
\begin{center}
   \includegraphics[width=0.7\linewidth]{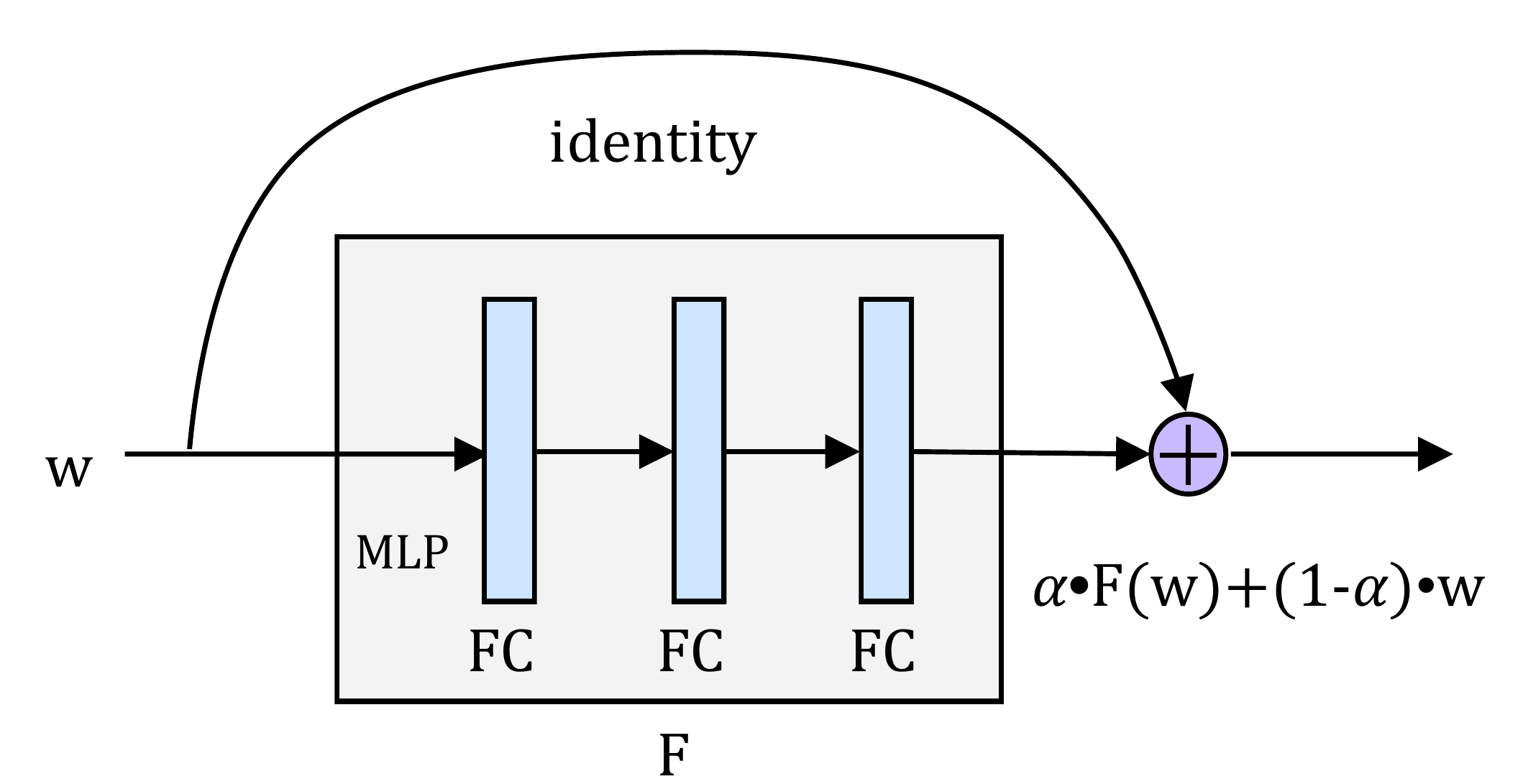}
\end{center}
   \caption{The proposed residual attention multi-layer perceptron (RA-MLP) layer.}
\label{fig:ra-mlp}
\end{figure}

\subsubsection{Attention based Style Manipulation}

\textbf{Residual Attention MLP.}
In this work, we achieve the non-linear style manipulation in the latent space by using multiple RA-MLP layers. As shown in Figure \ref{fig:ra-mlp}, RA-MLP utilizes a MLP and a shortcut connection to build a residual learning block \cite{he2016deep}, which not only enables the non-linear mapping in the latent space, but also provides the access to the original style information. Furthermore, we allow an attentive fusion scheme to balance the degree of semantic editing and original information preservation. Formally, we consider a building block defined as:
\begin{equation}
\label{eq:ra-mlp}
w^+ = \alpha * F(w) + (1 - \alpha) * w
\end{equation}
where $w$ and $w^+$ are the intermediate and extended latent code, respectively. The function $F$ represents the MLP layer to be learned. $\alpha$ is a learnable attention parameter with a range between 0 and 1. As a result, this design enables us to gradually edit the attribute of the input face. To the extreme, we can push $\alpha$ to zero if an identity mapping is desired, thus the input face will not be affected. Another benefit of this design is that it could be much easier to optimize a residual mapping than to directly optimize an unreferenced mapping.

\textbf{Attentive Attribute Editing.}
When there is more than one attribute, preforming a single attribute edit along all the style layers may affect another \cite{karras2019style,karras2020analyzing}. In order to achieve more disentangled editing, we provide additional constraints to enable the modified generator to perform attribute edits by attending to different style layers. Formally, let $a$ be an attention vector and its dimension is the same as the number of the layers in the synthesis network. Then the editing in a particular layer $j$ can be formulated as:
\begin{align}
\alpha &= softmax(a) \\
w^+_j &= \alpha_j * F_j(w) + (1 - \alpha_j) * w
\end{align}
where $\alpha_j$ is used to control the magnitude of changes in the $j^{th}$ layer. To ensure the attention vector to be sparse, we add an entropy penalty to the loss:
\begin{equation}
\small
\mathcal{L}_{entropy}(\alpha) = -\sum_{j}(\alpha_j\cdot\log(\alpha_j) + (1-\alpha_j)\cdot\log(1-\alpha_j))
\end{equation}
Thus, our full objective to edit a particular attribute $i$ is a combination of the cross-entropy loss and entropy penalty. We aim to solve:
\begin{equation}
\label{eq:loss}
h^*, \alpha^* = \arg\min_{h,\alpha}\mathcal{L}^{cls}_i(f,h,g,c) + \lambda \mathcal{L}_{entropy}(\alpha)
\end{equation}
where $\lambda$ controls the relative importance of the two components.

\textbf{Single Layer Manipulation.}
In order to get a more precise and disentangled edits, we only retain the strongest manipulation among all the style layers during test stage. As illustrated in Figure \ref{fig:overview}(b), the learned generator can adaptively manipulate a \textit{single} layer selected by the maximal attention value $\alpha_{\text{max}}$, and then continue the feedforward to produce an edited face image. In this way, we can easily achieve continuous control of facial attributes by gradually changing the attention value.

\section{Experiment}
\subsection{Experimental Settings}
\label{sec:details}
\textbf{Network Architecture.}
In this work, we conduct all the experiments on the recently proposed StyleGAN2 model \cite{karras2020analyzing}. We just add a style manipulation network with multiple RA-MLP layers in the generator and leave everything else (mapping network, adaptive instance normalization and synthesis network) untouched. Moreover, instead of retraining the model, we reuse the weights pretrained on face images. For face attribute estimation, we adopt the ResNet18 \cite{he2016deep} with a global average pooling as the shared encoder $E$, followed by 40 independent fully connected layers to estimate attributes. Our face attribute classifier is trained from scratch.

\textbf{Dataset.}
We adopt StyleGAN2 model pretrained on FFHQ dataset \cite{karras2020analyzing} for face editing. Additionally we conduct experiments to edit artistic faces by using StyleGAN2 trained on MetFaces dataset \cite{karras2020training}. We use CelebA dataset \cite{liu2015faceattributes} to train and evaluate our face attribute classifier. It consists of 202,599 real face images, each labeled with 5 landmark locations and 40 binary attributes annotations.

\textbf{Training Details.}
For face attribute estimation, we build a training dataset by cropping the aligned CelebA images to $128 \times 128$. We use Adam solver with a batch size of 64 for 30 epochs to train our attribute classifier. The initial learning rate is $3e^{-4}$ and decayed by a factor of 0.1 every 10 epochs. For semantic face editing, we randomly generate face images on the fly with the pretrained generator. We use SGD with a momentum weight of 0.9, learning rate of 0.001 and batch size of 10 to train the modified generator. We set $\lambda = 1$ in Equation \ref{eq:loss}. Note that the pretrained StyleGAN2 generator and our face attribute classifier are fixed during the whole training process.


\subsection{Attribute Classifier Pretraining}
\label{sec:classifier}
\begin{figure}[h]
\begin{center}
   \includegraphics[width=0.95\linewidth]{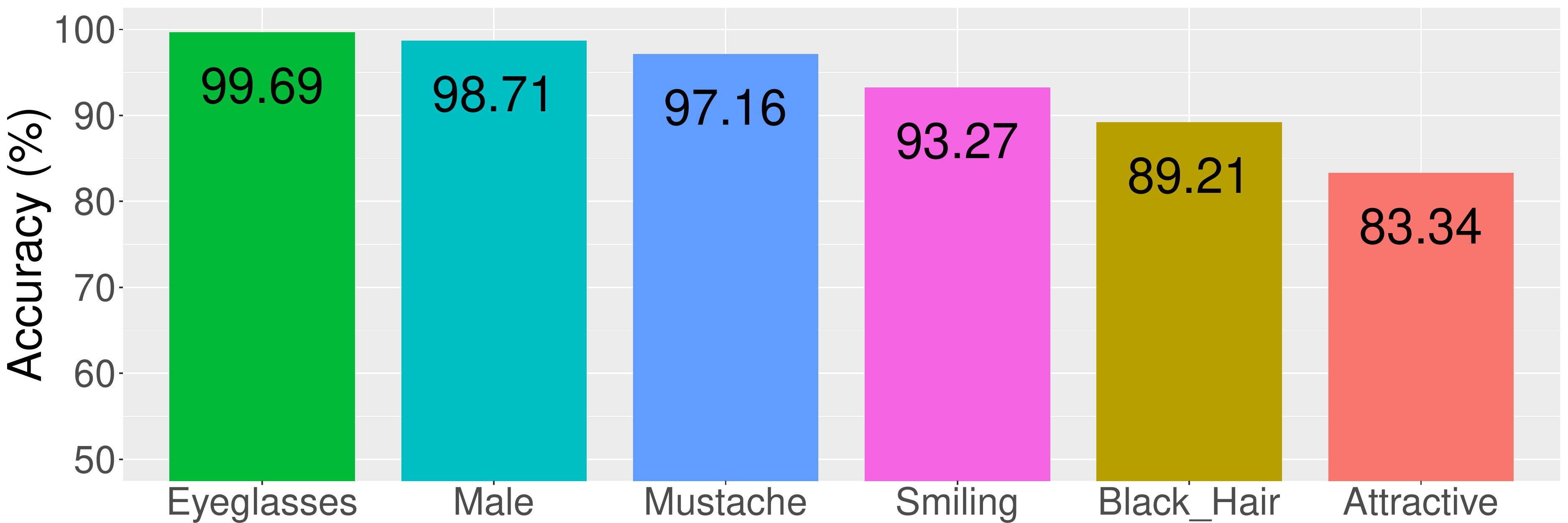}
\end{center}
   \caption{The results of face attribute estimation.}
\label{fig:prediction}
\end{figure}

We first present the results of multiple binary attributes estimation on CelebA dataset. As shown in Figure \ref{fig:prediction}, our attribute classifier can achieve 93.56\% average accuracy on 6 attributes considered in this work. This high performance suggests that the learned attribute classifier is able to capture the key features of different attributes. Thus, we can reuse the learned knowledge to guide the semantic face editing task.


\begin{figure}[!]
\begin{center}
   \includegraphics[width=0.93\linewidth]{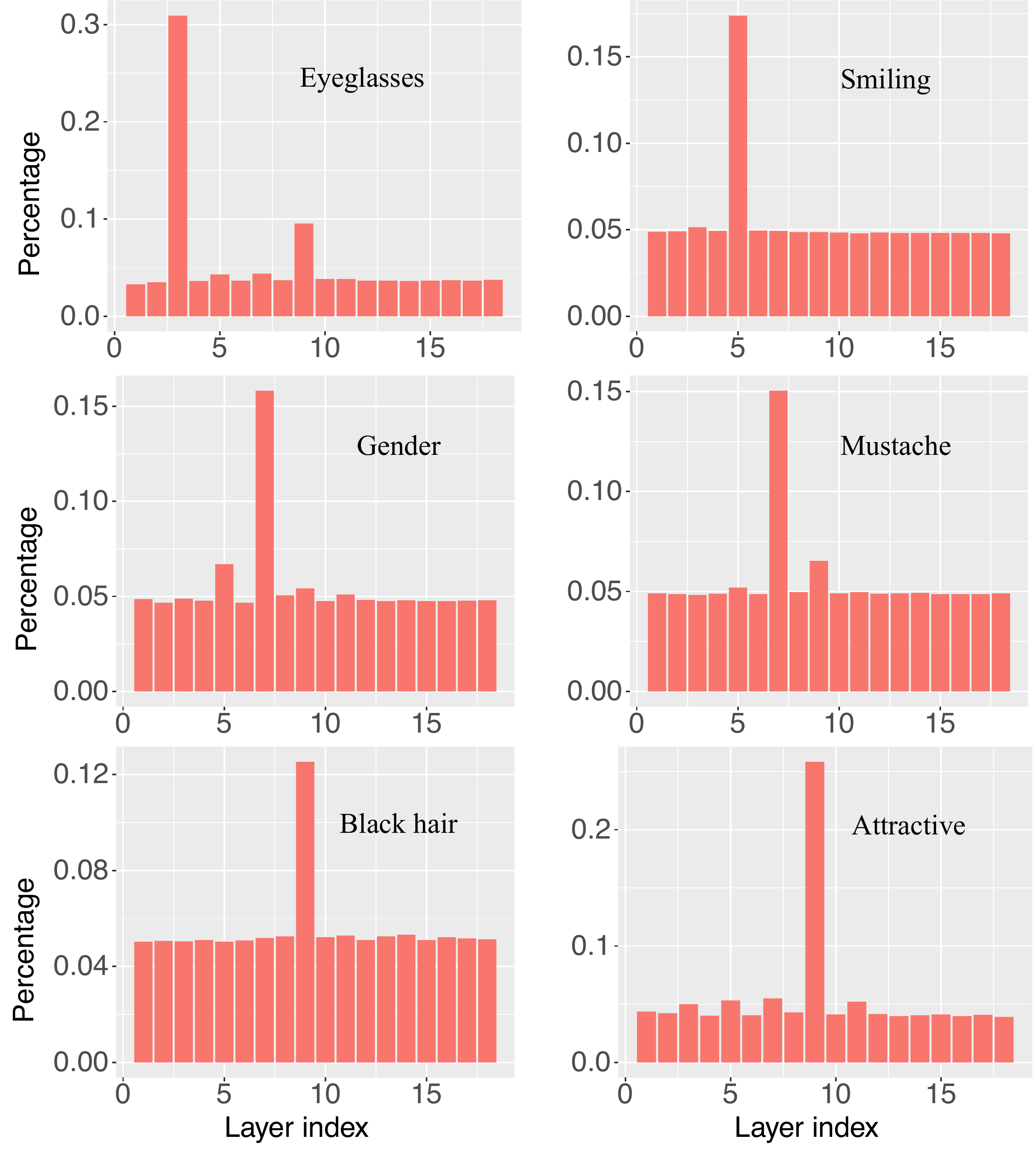}
\end{center}
   \caption{The results of learned attentions across different layers.}
\label{fig:attention}
\end{figure}

\begin{figure}[!]
\begin{center}
   \includegraphics[width=0.82\linewidth]{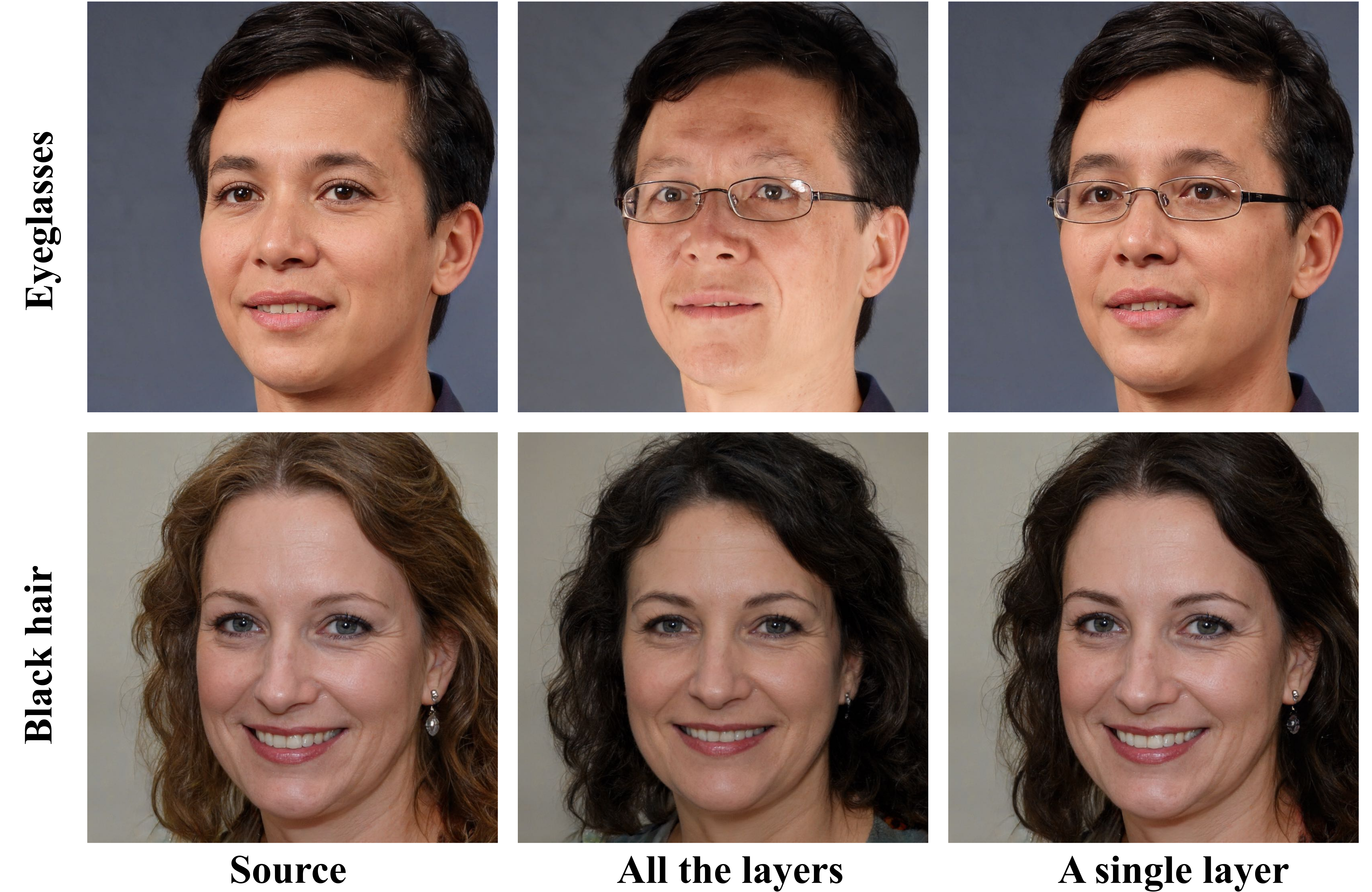}
\end{center}
   \caption{Effectiveness of single layer manipulation. From left to right: source images, edited results by using all the layers and a single layer.}
\label{fig:sparse}
\end{figure}

\subsection{Attention based Style Manipulation}
\textbf{Learning to Attend.}
In this part, we inspect the resulting attention component learned by our model. Figure \ref{fig:attention} shows the learned attentions across all the 18 layers in our generator for different attributes. We can easily tell that there exists one style layer that corresponds to a much bigger attention value $\alpha$ than others. In other words, our model can adaptively select a single layer to perform manipulation when editing a particular attribute. Additionally, we observe that high-level structured attributes (shape is changed) such as \textit{eyeglasses} (the $3^{rd}$ layer) and \textit{smiling} (the $5^{th}$ layer) are corresponding to first few layers while middle layers (such as the $9^{th}$ layer) can retain the overall face structure and mainly control the color scheme and textures like \textit{black hair} and \textit{attractive}. It implies that our generator is able to learn the underlying semantic representations in different layers, which is consistent with the observation from previous works \cite{karras2019style,shen2020interpreting}.

\textbf{Effectiveness of Single Layer Manipulation.}
We report experimental results to demonstrate the effectiveness by only manipulating a single style layer. Figure \ref{fig:sparse} shows the comparison of the proposed approach by using a single layer versus all the layers in the synthesis network. We can tell that better disentangled results can be achieved by manipulating a single style layer. By contrast, when performing edits of eyeglasses across multiple layers, other attributes such as expression, eyebrows and skin tone could be changed. Intuitively, we can get a better chance to preserve the identity information and achieve a more disentangled edit by avoiding the unnecessary changes of other layers.

\begin{figure}[]
\begin{center}
   \includegraphics[width=0.95\linewidth]{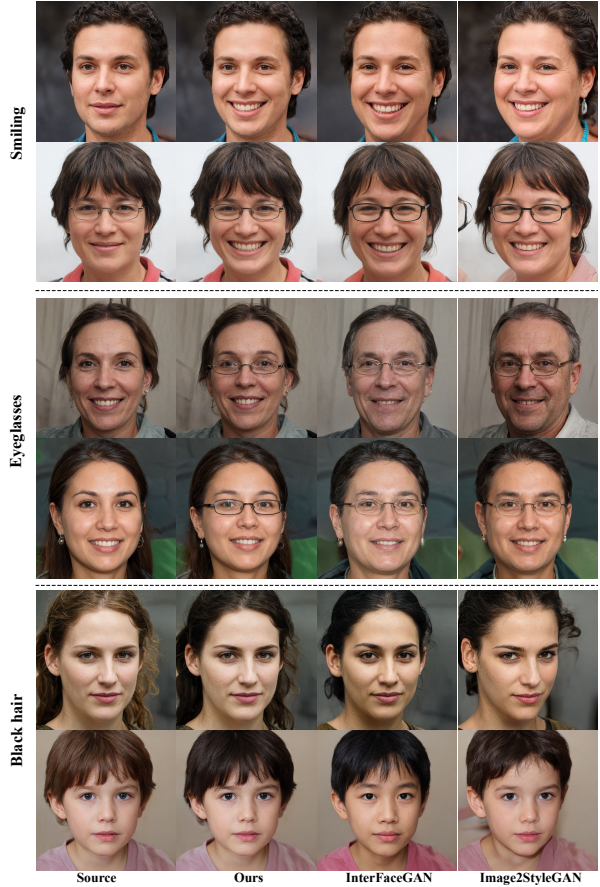}
\end{center}
   \caption{Visual comparison with other methods for semantic face editing. From left to right: source images, our method, InterFaceGAN \cite{shen2020interpreting} and Image2StyleGAN \cite{abdal2019image2stylegan}. From top to bottom we provide the edited results on \textit{smiling}, \textit{eyeglasses}, and \textit{black hair}.}
\label{fig:comparison}
\end{figure}

\begin{figure*}[!]
\begin{center}
   \includegraphics[width=0.87\linewidth]{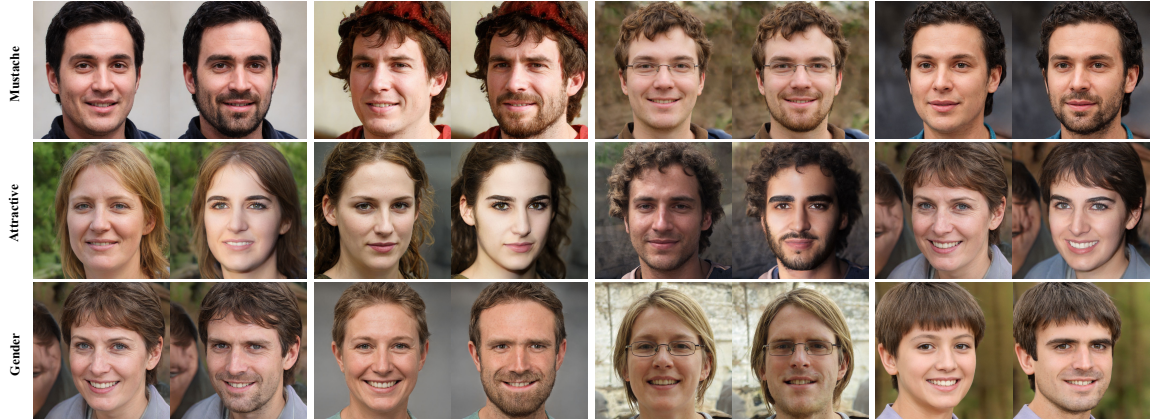}
\end{center}
   \caption{Visual results of our method to edit attributes like \textit{mustache}, \textit{attractive} and \textit{gender}.}
\label{fig:more}
\end{figure*}

\begin{figure*}[!]
\begin{center}
   \includegraphics[width=0.87\linewidth]{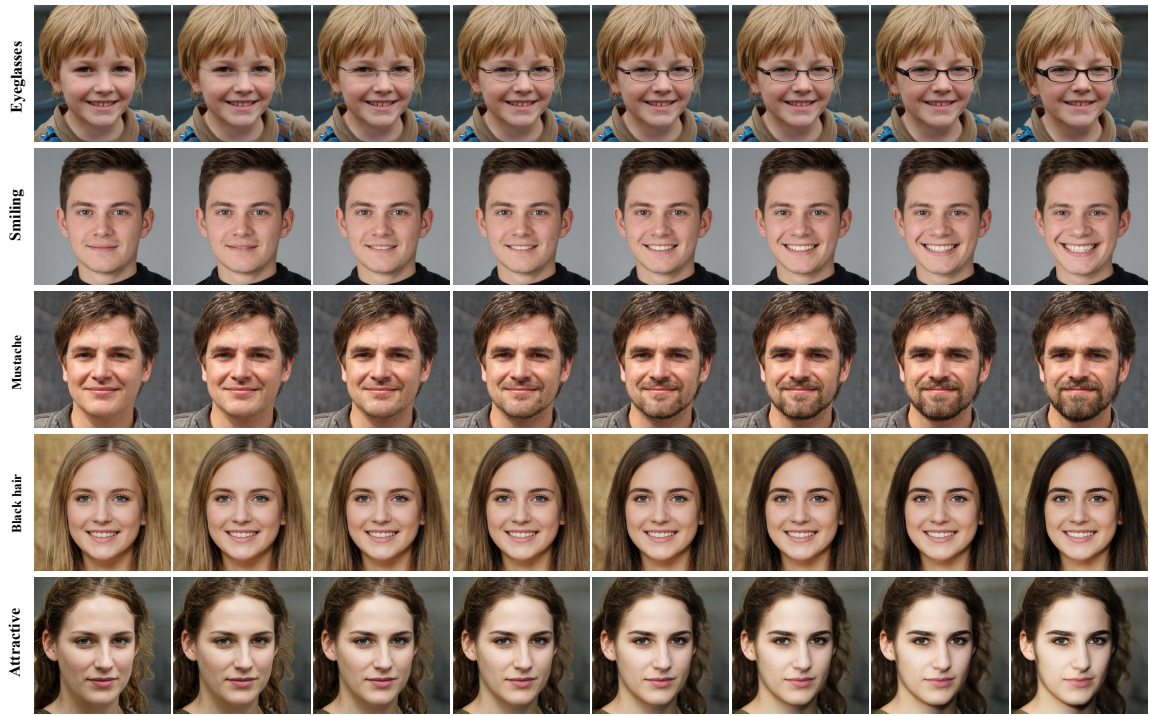}
\end{center}
   \caption{Visual results of continuous editing. Each row represents the edited results with linearly increased attention value $\alpha$.}
\label{fig:continuous}
\end{figure*}

\begin{figure}[!]
\begin{center}
   \includegraphics[width=0.94\linewidth]{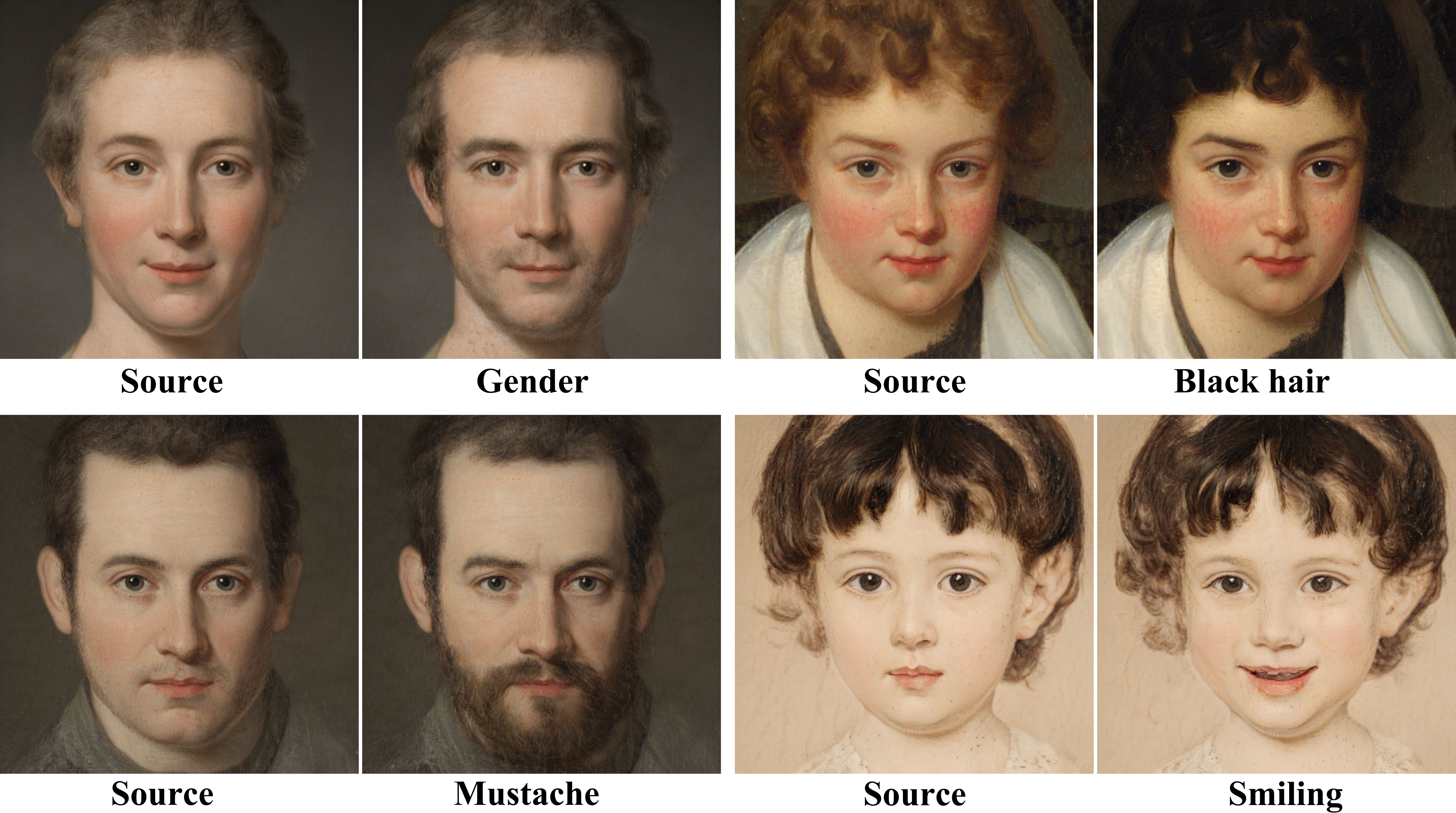}
\end{center}
   \caption{Visual results of our method for artistic face editing. We consider attributes like \textit{gender}, \textit{black hair}, \textit{mustache} and \textit{smiling}.
 }
\label{fig:artistic}
\end{figure}

\begin{figure}[!]
\begin{center}
   \includegraphics[width=0.95\linewidth]{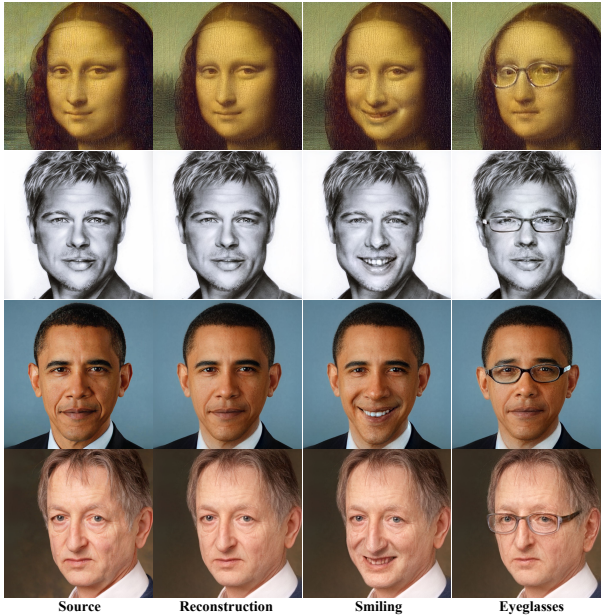}
\end{center}
   \caption{Visual results of our method for real face editing, including artistic painting, pencil drawing and portrait photography. From left to right: source images, reconstructed outputs via GAN inversion, the edited results on \textit{smiling} and \textit{eyeglasses}.}
\label{fig:real}
\end{figure}

\subsection{Semantic Face Editing}
\label{sec:editing}
In this section, we investigate the quality of semantic face editing achieved by our model. We first present the editing results on synthesized faces. Then more experiments are provided to demonstrate that our approach can also generalize well to real faces and artistic portraits.

\textbf{Synthesized Face Editing.}
As shown in Figure \ref{fig:abstract}, \ref{fig:comparison} and \ref{fig:more}, we can see that our approach is able to achieve impressive editing results along different attributes. In particular, our method can accurately edit both local attributes (smiling, eyeglasses, mustache and hair color) as well as global attributes (gender and attractive). For example, \textit{eyeglasses} and \textit{smiling} can be naturally added on both male and female faces. Particularly on \textit{smiling} attribute (Figure \ref{fig:abstract} and \ref{fig:comparison}), we can even observe realistic wrinkles under eyes caused by smiling. A \textit{female} can be also transferred to a \textit{male} face with distinct deep set eyes and visible mustache, while retaining overall facial appearance like background and color scheme (the third row in Figure \ref{fig:more}). Furthermore, we can even make a face more \textit{attractive} with dramatic masculine or feminine features. For instance, makeup styles can be transferred to generate a fair-skinned female face and beautiful eyelashes (the second row in Figure \ref{fig:more}). These high-quality editing results provide a strong evidence that our proposed GuidedStyle successfully transfers the knowledge learned by an attribute classifier to a GAN generator.

\textbf{Continuous Editing.}
It has been widely observed that linear interpolation in GAN latent space can lead to continuous changes of synthesized images. In this part, we provide more results to demonstrate that our approach also supports continuous editing for various facial attributes. Specifically, we can choose how much a given attribute is perceivable in the edited faces by using continuous attention values with Equation \ref{eq:ra-mlp}. The results are shown in Figure \ref{fig:continuous}. Each row shows the edited version of the same face with linearly increased attention value $\alpha$ and the left images are the sources. We can see that our model can make subtle changes to the attributes of the source faces, which are visually plausible with smooth transition. For example, it is possible to gradually change the expression to produce a smiling face, make the mustache of a man more noticeable and a female more attractive. Please also notice that other attributes are barely affected, demonstrating that our method can achieve disentangled and controllable face editing.

\textbf{Artistic Portrait Editing.}
In order to further verify that our GuidedStyle framework can effectively make use of the semantic knowledge learned by attribute classifiers, we provide additional experiments on artistic face editing. In our setting, we replace the StyleGAN2 generator with the one trained on artistic portraits (MetFaces dataset \cite{karras2020training}). We still use the same face attribute classifier to supervise the training of the new generator. Figure \ref{fig:abstract} (the second row) and Figure \ref{fig:artistic} show the edited results of artistic portraits on different attributes. We notice that in spite of the large domain gap between the two databases, the edits are still of very high quality. Take the second row in Figure \ref{fig:abstract} as an example, we can change the expression and even add eyeglasses and mustache to the same face while keeping other details unchanged. It is also possible to generate the corresponding male version of an artistic female (see Figure \ref{fig:artistic}).









\textbf{Real Face Editing.}
Besides performing editing on synthesized faces, we further apply our approach to real face editing. In particular, we start with a real image and employ the projection method in \cite{karras2020analyzing} to extract the corresponding latent code, which is then fed to our generator to produce the edited faces. We show the visual results in Figure \ref{fig:real}. One can see that our method demonstrates high quality edited results on different types of real faces, including \textit{artistic painting}, \textit{pencil drawing} and \textit{portrait photography} with different poses. In the fourth column, for instance, realistic eyeglasses can be added without affecting other attributes. We would also like to note that the eyeglasses are not just randomly generated. Instead, the \textit{selected} eyeglass frames fit all the faces properly in terms of the shape and color scheme. And these results sheds light on the superior generalization ability of our approach.

\begin{table}[]
\tabcolsep=0.2cm
\centering
\caption{Quantitative results measured by using different metrics. $\downarrow$ indicates that lower is better, and $\uparrow$ indicates that higher is better.}
\scalebox{0.85}{
\begin{tabular}{c|c|c|c|c}
\hline
\multirow{2}{*}{Method} & \multicolumn{2}{c|}{Image quality} & \multicolumn{2}{c}{Identity preservation} \\ \cline{2-5}
                        & ~~~~FID$\downarrow$~~~~             & ~~SWD$\downarrow$~~              & ~~~~CS$\uparrow$~~~~                 & ED$\downarrow$                  \\ \hline
InterfaceGAN            & 49.90           & 262.44           & 0.60                 & 0.87                \\
Image2StyleGAN          & 53.08           & 246.16           & 0.56                 & 0.90                \\
Ours                    & \textbf{41.79}  & \textbf{218.71}  & \textbf{0.64}        & \textbf{0.79}       \\ \hline
\end{tabular}}
\label{tab:comparison}
\end{table}

\subsection{Comparison with other Methods}
\label{sec:comparison}

\textbf{Qualitative Comparison.}
Figure \ref{fig:comparison} shows the qualitative comparison of our method with two recently proposed editing techniques, InterFaceGAN \cite{shen2020interpreting} and Image2StyleGAN \cite{abdal2019image2stylegan}. We can tell that our method can achieve better results, and demonstrate more disentangled edits along various attributes, including \textit{smiling}, \textit{eyeglasses} and \textit{black hair}. For example, when editing eyeglasses attribute, the input female becomes male with changed background when using InterFaceGAN and Image2StyleGAN. In the case of hair color editing, InterFaceGAN also tends to darken the skin when producing black hair, and there exist noticeable changes of hair style and face pose. By contrast, our method yields a more controllable edits and demonstrates superior identity preservation.

\textbf{Quantitative Comparison.}
The superiority of our method can be also validated by the quantitative evaluation as shown in Table \ref{tab:comparison}, where our approach outperforms others by a large margin. Specifically, we generate 1,000 face images from StyleGAN2 and perform edits along \textit{smiling}, \textit{eyeglasses}, \textit{mustache}, \textit{black hair} and \textit{attractive}. Then four metrics are calculated for evaluation, including Fr\'{e}chet Inception Distance (FID) \cite{heusel2017gans}, Sliced Wasserstein Distance (SWD) \cite{rabin2011wasserstein}, Cosine Similarity (CS) and Euclidean Distance (ED). CS and ED are calculated by using the image embeddings extracted from face recognition model FaceNet \cite{schroff2015facenet}. FID and SWD are used to measure the statistical similarity between edited faces and the originals, while CS and ED are used to quantify the identity preservation.






\section{Conclusion}
In this work, we introduce GuidedStyle to explore the latent semantics in GANs for high-level face editing. Our approach features a novel learning framework that leverages the knowledge learned from face attribute classifiers to guide the image generation process of StyleGAN. In addition, our method allows a sparse attention mechanism to select a single layer of the generator for semantic face editing. As a result, we are able to disentangle a variety of semantics and achieve precise edits along different facial attributes. Extensive experiments demonstrate the superior performance of our method over previous works. Moreover, our editing model generalizes well to artistic portraits and real face images.

{\small
\bibliographystyle{ieee_fullname}
\bibliography{egbib}
}

\end{document}